\documentclass[journal]{IEEEtran}
\ifCLASSINFOpdf
\else
\fi
\usepackage{amssymb,hyperref}
\usepackage{amsmath,amsfonts,paralist,booktabs,booktabs,cleveref,ntheorem}
\usepackage{graphicx,multirow,cite}

\newtheorem{remark}{Remark}
\begin{document}
%
\title{Maximum Correntropy Unscented Filter}
%
%
%

\author{Xi~Liu,
        Badong~Chen,
        Bin~Xu,
        Zongze~Wu
        and~Paul~Honeine
\thanks{This work was supported by 973 Program (No. 2015CB351703) and the National Natural Science Foundation of China (No. 61372152).}
\thanks{X. Liu and B. Chen are with the School of Electronic and Information Engineering, Xi'an Jiaotong University, Xi'an, China. (e-mail: lx1102@stu.xjtu.edu.cn ; chenbd@mail.xjtu.edu.cn).}
\thanks{B. Xu is with the School of Automation, Northwestern Polytechnical University, Xi'an, China. (e-mail: smileface.binxu@gmail.com).}
\thanks{Z. Wu is with the School of Electronic and Information Engineering, South China University of Technology, Guangzhou, China. (e-mail: zzwu@scut.edu.cn).}
\thanks{P. Honeine is with the Normandie Univ, UNIROUEN, UNIHAVRE, INSA Rouen, LITIS, Rouen, France. (email: paul.honeine@univ-rouen.fr).}}

\maketitle

\begin{abstract}
The unscented transformation (UT) is an efficient method to solve the state estimation problem for a non-linear dynamic system, utilizing a derivative-free higher-order approximation by approximating a Gaussian distribution rather than approximating a non-linear function. Applying the UT to a Kalman filter type estimator leads to the well-known unscented Kalman filter (UKF). Although the UKF works very well in Gaussian noises, its performance may deteriorate significantly when the noises are non-Gaussian, especially when the system is disturbed by some heavy-tailed impulsive noises. To improve the robustness of the UKF against impulsive noises, a new filter for nonlinear systems is proposed in this work, namely the maximum correntropy unscented filter (MCUF). In MCUF, the UT is applied to obtain the prior estimates of the state and covariance matrix, and a robust statistical linearization regression based on the maximum correntropy criterion (MCC) is then used to obtain the posterior estimates of the state and covariance. The satisfying performance of the new algorithm is confirmed by two illustrative examples.
\end{abstract}

\begin{IEEEkeywords}
Unscented Kalman Filter (UKF), Unscented Transformation (UT), Maximum Correntropy Criterion (MCC).
\end{IEEEkeywords}

%
\IEEEpeerreviewmaketitle

\section{Introduction}
%
%
%
%
\IEEEPARstart{E}{stimation} problem plays a key role in many fields, including communication, navigation, signal processing, optimal control and so on \cite{Li2013unscented,Li2010fault,Dash2010adaptive,Partovibakhsh2014adaptive}. The Kalman filter (KF) assists in obtaining accurate state estimation for a linear dynamic system, which provides an optimal recursive solution under minimum mean square error (MMSE) criterion \cite{Kalman1960a,Bryson1975applied,Nahi1969estimation}. Nevertheless, most practical systems are inherently nonlinear, and it is not easy to implement an optimal filter for nonlinear systems. To solve the nonlinear filtering problem, so far many sub-optimal nonlinear extensions of the KF have been developed by using some approximations, among which the extended Kalman filter (EKF) \cite{Anderson1979optimal} and unscented Kalman filter (UKF) \cite{Julier2000a} are two widely used ones. As a popular nonlinear extension of KF, the EKF approximates the nonlinear system by its first order linearization and uses the original KF on this approximation. However, the crude approximation may lead to divergence of the filter when the function is highly non-linear. Moreover, the cumbersome derivation of the Jacobian matrices often leads to the implementation difficulties. The UKF is an alternative to the EKF, which approximates the probability distribution of the state by a set of deterministically chosen sigma points and propagates the distribution though the non-linear equations. The UKF does not need to calculate the Jacobian matrices and can obtain a better performance than the EKF. However, the UKF may perform poorly when the system is disturbed by some heavy-tailed non-Gaussian noises, which occur frequently in many real-world applications of engineering. The main reason for this is that the UKF is based on the MMSE criterion and thus exhibits sensitivity to heavy-tailed noises \cite{Schick1994robust}. Some methods have been proposed to cope with this problem in the literatures. In particular, the Huber's generalized maximum likelihood methodology is an important one \cite{Huber1964robust,Hawary1995robust,Wang2010huber,Karlgaard2007huber}, which is a combined minimum ${\ell}_1$ and ${\ell}_2$ norm estimation method. Furthermore, the statistical linear method instead of the first order linearization was introduced in \cite{Karlgaard2007huber}.

Besides Huber's robust statistics, information theoretic quantities (e.g. entropy, mutual information, divergence, etc.) can also be used as a robust cost for estimation problems \cite{Principe2010information,Chen2013system}. As a localized similarity measure in information theoretic learning (ITL), the correntropy has recently been successfully applied in robust machine learning and non-Gaussian signal processing \cite{Liu2007correntropy,Chen2014steady,Zhao2011kernel,Singh2009using,Chen2012maximum,Wang2015robust,Chen2012recursive,He2014robust,Chen2016generalized,He2011maximum,He2011robust,Xu2008a,Bessa2009entropy,Shi2014convex}. The adaptive filtering algorithms under the maximum correntropy criterion (MCC) can achieve excellent performance in heavy-tailed non-Gaussian noises \cite{Principe2010information,Chen2014steady,Zhao2011kernel,Singh2009using,Wang2015robust,Chen2016generalized,Shi2014convex}. In particular, in a recent work \cite{Chen2015maximum}, a linear Kalman type filter has been developed under MCC, which can outperform the original KF significantly especially when the disturbance noises are impulsive.

The goal of this paper is to develop an unscented non-linear Kalman type filter based on the MCC, called the maximum correntropy unscented filter (MCUF). In the MCUF, the unscented transformation (UT) is applied to get a prior estimation of the state and covariance matrix, and a statistical linearization regression model based on MCC is used to obtain the posterior state and covariance. The new filter adopts the UT and statistical linear approximation instead of the first order approximation as in EKF to approximate the nonlinearity, and uses the MCC instead of the MMSE to cope with the non-Gaussianity, and hence can achieve desirable performance in highly nonlinear and non-Gaussian conditions. Moreover, the proposed MCUF is suitable for online implementation on account of the retained recursive structure.

The rest of the paper is organized as follows. In Section \ref{sec:2}, we briefly introduce the MCC. In Section \ref{sec:3}, we derive the MCUF algorithm. In  Section \ref{sec:4}, we present two illustrative examples and show the desirable performance of the proposed algorithm. Finally, Section \ref{sec:5} concludes this paper.


\section{Maximum correntropy criterion}
\label{sec:2}
Correntropy is a generalized similarity measure between two random variables. Given two random variables $X,Y \in \mathbb{R}$ with joint distribution function ${\operatorname{F}_{XY}}( x,y )$, the correntropy is defined by
\begin{equation}\label{equ:1}
 V(X,Y) = {\operatorname E}\left[ {\kappa (X,Y)} \right] = \int {\kappa (x,y)d{\operatorname{F}_{XY}}} (x,y)
\end{equation}
where $\operatorname{E}$ denotes the expectation operator, and $\kappa ( \cdot , \cdot )$ is a shift-invariant Mercer kernel. In this work, without mentioned otherwise, the used kernel function is the Gaussian kernel:
\begin{equation}\label{equ:2}
 \kappa (x,y) = {{\operatorname G}_\sigma }(e) = \exp \left( { - \frac{{{e^2}}}{{2{\sigma ^2}}}} \right)
\end{equation}
where $e= x - y$, and $\sigma  > 0$ stands for the kernel bandwidth.

In many practical situations, we have only a set of finite data and the joint distribution ${\operatorname F}_{XY}$ is unknown. In these cases, one can estimate the correntropy using a sample mean estimator:
\begin{equation}\label{equ:3}
 \widehat V(X,Y) = \frac{1}{N}\sum\limits_{i = 1}^N {{{\operatorname G}_\sigma }} \left( {e(i)} \right)
\end{equation}
where $e(i) = x(i) - y(i)$, with $\left\{ {x(i),y(i)} \right\}_{i = 1}^N$ being $N$ samples drawn from ${\operatorname F}_{XY}$.

Using the Taylor series expansion for the Gaussian kernel yields
\begin{equation}\label{equ:4}
 V( X,Y ) = \sum\limits_{n = 0}^\infty  {\frac{{{{\left( { - 1} \right)}^n}}}{{{2^n}{\sigma ^{2n}}n!}}} {\rm E}\left[ {{{(X - Y)}^{2n}}} \right]
\end{equation}
Thus, the correntropy is a weighted sum of all even order moments of the error variable $X - Y$. The kernel bandwidth appears as a parameter to weight the second order and higher order moments. With a very large kernel bandwidth (compared to the dynamic range of the data), the correntropy will be dominated by the second order moment.

Suppose our goal is to learn a parameter vector $W$ of an adaptive model, and let $x(i)$ and $y(i)$ denote, respectively, the model output and the desired response. The MCC based learning can be formulated as solving the following optimization problem:
\begin{equation}\label{equ:6}
 \widehat W = \mathop {\arg \max }\limits_{W \in \Omega } \frac{1}{N}\sum\limits_{i = 1}^N {{{\operatorname G}_\sigma }} \left( {e(i)} \right)
\end{equation}
where $\widehat W$ denotes the optimal solution, and $\Omega$ denotes a feasible set of the parameter.

\section{Maximum correntropy unscented filter}
\label{sec:3}

In this section, we propose to combine the MCC and a statistical linear regression model together to derive a novel non-linear filter, which can perform very well in non-Gaussian noises, since correntropy embraces second and higher order moments of the error.

Let's consider a nonlinear system described by the following equations:
\begin{equation}\label{equ:6}
{\mathbf{x}}(k) = {\mathop{\operatorname f}\nolimits} \left( {k - 1,{\mathbf{x}}(k - 1)} \right) + {\mathbf{q}}(k - 1),
\end{equation}
\begin{equation}\label{equ:7}
{\mathbf{y}}(k) = {\mathop{\operatorname h}\nolimits} \left( {k,{\mathbf{x}}(k)} \right) + {\mathbf{r}}(k).
\end{equation}
where ${\mathbf{x}}(k) \in {\mathbb{R}}{^n}$ denotes a $n$-dimensional state vector at time step $k$, ${\mathbf{y}}(k) \in {\mathbb{R}}{^m}$ represents an $m$-dimensional measurement vector, $\operatorname f$ is a nonlinear system function, and $\operatorname h$ is a nonlinear measurement function and both are assumed to be continuously differentiable. The process noise ${\mathbf{q}}(k - 1)$ and measurement noise ${\mathbf{r}}(k)$ are generally assumed to meet the independence assumption with zero mean and covariance matrices
\begin{equation}\label{equ:8}
\resizebox{0.9\hsize}{!}
 {${\operatorname E}\left[ {{\mathbf{q}}(k - 1){{\mathbf{q}}^T}(k - 1)} \right] = {\mathbf{Q}}(k - 1), \ {\operatorname E}\left[ {{\mathbf{r}}(k){{\mathbf{r}}^T}(k)} \right] = {\mathbf{R}}(k)$}
\end{equation}
Similar to other Kalman type filters, the MCUF also includes two steps, namely the time update and measurement update:

\subsection{Time update}

A set of $2n+1$ samples, also called sigma points, are generated from the estimated state $\widehat {\bf{x}}(k - 1|k - 1)$ and covariance matrix ${\bf{P}}(k - 1|k - 1)$ at the last time step $k-1$:
\begin{equation}\label{equ:9}
\resizebox{.91\hsize}{!}
{$\begin{split}
{\chi ^0}(k - 1|k - 1) = \ &\widehat {\bf{x}}(k - 1|k - 1),\\
{\chi ^i}(k - 1|k - 1) = \ &\widehat {\bf{x}}(k - 1|k - 1)\\
\ +& {\left( {\sqrt {(n + \lambda ){\bf{P}}(k - 1|k - 1)} } \right)_i},\ {\text{for}}\ i = 1 \ldots n,\\
{\chi ^i}(k - 1|k - 1) = \ &\widehat {\bf{x}}(k - 1|k - 1)\\
\ -& {\left( {\sqrt {(n + \lambda ){\bf{P}}(k - 1|k - 1)} } \right)_{i-n}},\ {\text{for}}\ i = n + 1 \ldots 2n.\\
\end{split}$}
\end{equation}
where ${\left( {\sqrt {(n + \lambda ){\bf{P}}(k - 1|k - 1)} } \right)_i}$ is the $i$th column of the matrix square root of $(n + \lambda ){\bf{P}}(k - 1|k - 1)$, with $n$ being the state dimension and $\lambda$ being a composite scaling factor, given by
\begin{equation}\label{equ:10}
\lambda  = {\alpha ^2}(n + \phi ) - n
\end{equation}
where $\alpha$ determines the spread of the sigma points, usually selected as a small positive number, and $\phi$ is a parameter that is often set to $3-n$.

The transformed points are then given through the process equation:
\begin{equation}\label{equ:11}
{\chi ^i}^ * (k|k - 1) = {\mathop{\operatorname f}\nolimits}\left( {k - 1,{\chi ^i}(k - 1|k - 1)} \right),\ {\text{for}}\ i = 0 \ldots 2n
\end{equation}
The prior state mean and covariance matrix are thus estimated by
\begin{equation}\label{equ:12}
\widehat {\bf{x}}(k|k - 1) = \sum\limits_{i = 0}^{2n} {w_m^i{\chi ^i}^ * (k|k - 1)} ,
\end{equation}
\begin{equation}\label{equ:13}
\resizebox{.89\hsize}{!}
{$\begin{split}
{\bf{P}}(k|k - 1) = &\sum\limits_{i = 0}^{2n} {w_c^i\left[ {{\chi ^i}^ * (k|k - 1) - \widehat {\bf{x}}(k|k - 1)} \right]} \\
&\times {\left[ {{\chi ^i}^ * (k|k - 1) - \widehat {\bf{x}}(k|k - 1)} \right]^T} + {\bf{Q}}(k - 1).
\end{split}$}
\end{equation}
in which the corresponding weights of the state and covariance matrix are
\begin{equation}\label{equ:14}
\begin{array}{l}
w_m^0 = \dfrac{\lambda }{{(n + \lambda )}}, \\[10pt]
w_c^0 = \dfrac{\lambda }{{(n + \lambda )}} + (1 - {\alpha ^2} + \beta ), \\
w_m^i = w_c^i = \dfrac{1}{{2(n + \lambda )}},\ {\text{for}}\ i = 1 \ldots 2n.
\end{array}
\end{equation}
where $\beta$ is a parameter related to the prior knowledge of the distribution of ${\bf{x}}(k)$ and is set to $2$ in the case of the Gaussian distribution.

\subsection{Measurement update}

Similarly, a set of $2n+1$ sigma points are generated from the prior state mean and covariance matrix
\begin{equation}\label{equ:15}
\resizebox{.89\hsize}{!}
{$\begin{split}
{\chi ^0}(k|k - 1) = \ &\widehat {\bf{x}}(k|k - 1),\\
{\chi ^i}(k|k - 1) = \ &\widehat {\bf{x}}(k|k - 1)\\
\ +& {\left( {\sqrt {(n + \lambda ){\bf{P}}(k|k - 1)} } \right)_i},\ {\text{for}}\ i = 1 \ldots n,\\
{\chi ^i}(k|k - 1) = \ &\widehat {\bf{x}}(k|k - 1)\\
\ -& {\left( {\sqrt {(n + \lambda ){\bf{P}}(k|k - 1)} } \right)_{i-n}},\ {\text{for}}\ i = n + 1 \ldots 2n.\\
\end{split}$}
\end{equation}
These points are transformed through the process equation as
\begin{equation}\label{equ:16}
{\gamma ^i}(k) = {\mathop{\operatorname h}\nolimits} (k,{\chi ^i}(k|k - 1)),\ {\text{for}}\ i = 0 \ldots 2n
\end{equation}
The prior measurement mean can then be obtained as
\begin{equation}\label{equ:17}
\widehat {\bf{y}}(k) = \sum\limits_{i = 0}^{2n} {w_m^i} {\gamma ^i}(k),
\end{equation}
Further, the state-measurement cross-covariance matrix is given by
\begin{equation}\label{equ:18}
\resizebox{.89\hsize}{!}
{${{\bf{P}}_{{\bf{xy}}}}(k) = \scalebox{1.2}{$\sum\limits_{i = 0}^{2n}$} {w_c^i} \left[ {{\chi ^i}(k|k - 1) - \widehat {\bf{x}}(k|k - 1)} \right]{\left[ {{\gamma ^i}(k) - \widehat {\bf{y}}(k)} \right]^T}.$}
\end{equation}

Next, we apply a statistical linear regression model based on the MCC to accomplish the measurement update. First, we formulate the regression model. We denote the prior estimation error of the state by
\begin{equation}\label{equ:19}
\eta ({\bf{x}}(k)) = {\bf{x}}(k) - \widehat {\bf{x}}(k|k - 1)
\end{equation}
and define the measurement slope matrix as
\begin{equation}\label{equ:20}
{\bf{H}}(k) = {\left( {{\bf{P}}^{ - 1}{(k|k - 1)}{{\bf{P}}_{{\bf{xy}}}}(k)} \right)^T}
\end{equation}
Then the measurement equation (\ref{equ:7}) can be approximated by \cite{Wang2010huber}
\begin{equation}\label{equ:21}
{\bf{y}}(k) \approx \widehat {\bf{y}}(k) + {\bf{H}}(k)({\bf{x}}(k) - \widehat {\bf{x}}(k|k - 1)) + {\bf{r}}(k)
\end{equation}

Combining (\ref{equ:12}) (\ref{equ:17}) and (\ref{equ:21}), we obtain the following statistical linear regression model:
\begin{equation}\label{equ:22}
\left[ {\begin{array}{*{20}{c}}
{\widehat {\bf{x}}(k|k - 1)}\\
{{\bf{y}}(k) - \widehat {\bf{y}}(k) + {\bf{H}}(k)\widehat {\bf{x}}(k|k - 1)}
\end{array}} \right] = \left[ {\begin{array}{*{20}{c}}
{\bf{I}}\\
{{\bf{H}}(k)}
\end{array}} \right]{\bf{x}}(k) + {\bf{\xi }}(k)
\end{equation}
where ${\bf{\xi }}(k)$ is
\begin{equation*}
{\bf{\xi }}(k) = \left[ {\begin{array}{*{20}{c}}
{\eta ({\bf{x}}(k))}\\
{{\bf{r}}(k)}
\end{array}} \right]
\end{equation*}
with
\begin{equation}\label{equ:23}
\begin{split}
{\bf{\Xi }}(k) = \ &{\operatorname E}\left[ {{\bf{\xi }}(k){{\bf{\xi }}^T}(k)} \right]\\
 = \ &\left[ {\begin{array}{*{20}{c}}
{{\bf{P}}(k|k - 1)}&0\\
0&{{\bf{R}}(k)}
\end{array}} \right]\\
 = \ &\left[ {\begin{array}{*{20}{c}}
{{{\bf{S}}_p}(k|k - 1){\bf{S}}_p^T(k|k - 1)}&0\\
0&{{{\bf{S}}_r}(k){\bf{S}}_r^T(k)}
\end{array}} \right]\\
 = \ &{\bf{S}}(k){{\bf{S}}^T}(k)
\end{split}
\end{equation}
Here, ${\bf{S}}(k)$ can be obtained by the Cholesky decomposition of ${\bf{\Xi }}(k)$. Left multiplying both sides of (\ref{equ:22}) by ${{\bf{S}}^{ - 1}}(k)$, the statistical regression model is transformed to
\begin{equation}\label{equ:24}
{\bf{D}}(k) = {\bf{W}}(k){\bf{x}}(k) + {\bf{e}}(k)
\end{equation}
where
\begin{equation*}
\begin{split}
{\bf{D}}(k) = \ &{{\bf{S}}^{ - 1}}(k)\left[ {\begin{array}{*{20}{c}}
{\widehat {\bf{x}}(k|k - 1)}\\
{{\bf{y}}(k) - \widehat {\bf{y}}(k) + {\bf{H}}(k)\widehat {\bf{x}}(k|k - 1)}
\end{array}} \right], \\
{\bf{W}}(k) = \ &{{\bf{S}}^{ - 1}}(k)\left[ {\begin{array}{*{20}{c}}
{\bf{I}}\\
{{\bf{H}}(k)}
\end{array}} \right], \\
{\bf{e}}(k) = \ &{{\bf{S}}^{ - 1}}(k){\bf{\xi }}(k).\\
\end{split}
\end{equation*}
It can be seen that ${\operatorname E}\left[ {{\bf{e}}(k){{\bf{e}}^T}(k)} \right] = {\bf{I}} $.

We are now in a position to define a cost function based on the MCC:
\begin{equation}\label{equ:25}
{J_L}\left( {{\bf{x}}(k)} \right) = \sum\limits_{i = 1}^L {{{\operatorname G}_\sigma }\left( {{d_i}(k) - {{\bf{w}}_i}(k){\bf{x}}(k)} \right)}
\end{equation}
where ${d_i}(k)$ is the $i$-th element of ${\bf{D}}(k)$, ${{\bf{w}}_i}(k)$ is the $i$-th row of ${\bf{W}}(k)$, and $L=n+m$ is the dimension of ${\bf{D}}(k)$.
Under the MCC, the optimal estimate of ${\bf{x}}(k)$ arises from the following optimization:
\begin{equation}\label{equ:26}
\widehat {\bf{x}}(k) = \arg \mathop {\max }\limits_{{\bf{x}}(k)} {J_L}\left( {{\bf{x}}(k)} \right) = \arg \mathop {\max }\limits_{{\bf{x}}(k)} \sum\limits_{i = 1}^L {{{\operatorname G}_\sigma }} \left( {{e_i}(k)} \right)
\end{equation}
where ${e_i}(k)$ is the $i$-th element of ${\bf{e}}(k)$ given by
\begin{equation}\label{equ:27}
{e_i}(k) = {d_i}(k) - {{\bf{w}}_i}(k){\bf{x}}(k)
\end{equation}
The optimal solution of ${\bf{x}}(k)$ can be solved through
\begin{equation}\label{equ:28}
\frac{{\partial {J_L}\left( {{\bf{x}}(k)} \right)}}{{\partial {\bf{x}}(k)}} = 0
\end{equation}
It follows easily that
\begin{equation}\label{equ:29}
\begin{split}
{\bf{x}}(k) = \ &{\left( {\sum\limits_{i = 1}^L {\left( {{{\operatorname G}_\sigma }\left( {{e_i}(k)} \right){\bf{w}}_i^T(k){{\bf{w}}_i}(k)} \right)} } \right)^{ - 1}} \times \\
&\left( {\sum\limits_{i = 1}^L {\left( {{{\operatorname G}_\sigma }\left( {{e_i}(k)} \right){\bf{w}}_i^T(k){d_i}(k)} \right)} } \right)
\end{split}
\end{equation}
Since ${e_i}(k) = {d_i}(k) - {{\bf{w}}_i}(k){\bf{x}}(k)$, the equation (\ref{equ:29}) is actually a fixed-point equation with respect to ${\bf{x}}(k)$ and can be rewritten as
\begin{equation}\label{equ:30}
{\bf{x}}(k) = \operatorname g\left( {{\bf{x}}(k)} \right)
\end{equation}
Hence, a fixed-point iterative algorithm can be obtained as \cite{Agarwal2001fixed,Singh2010aclosed,Chen2015convergence}
\begin{equation}\label{equ:31}
\widehat {\bf{x}}{(k)_{t + 1}} = \operatorname g\left( {\widehat {\bf{x}}{{(k)}_t}} \right)
\end{equation}
with $\widehat {\bf{x}}{(k)_t}$ being the estimated state $\widehat {\bf{x}}(k)$ at the $t$-th fixed-point iteration .

The fixed-point equation (\ref{equ:29}) in matrix form can also be expressed as
\begin{equation}\label{equ:32}
{\bf{x}}(k) = {\left( {{{\bf{W}}^T}(k){\bf{C}}(k){\bf{W}}(k)} \right)^{ - 1}}{{\bf{W}}^T}(k){\bf{C}}(k){\bf{D}}(k)
\end{equation}
where ${\bf{C}}(k) = \left[ {\begin{array}{*{20}{c}}
{{{\bf{C}}_x}(k)}&0\\
0&{{{\bf{C}}_y}(k)}
\end{array}} \right]$, with\\ ${{\bf{C}}_x}(k) = diag\left( {{{\operatorname G}_\sigma }\left( {{e_1}(k)} \right),...,{{\operatorname G}_\sigma }\left( {{e_n}(k)} \right)} \right)$,\\ ${{\bf{C}}_y}(k) = diag\left( {{{\operatorname G}_\sigma }\left( {{e_{n + 1}}(k)} \right),...,{{\operatorname G}_{n + m}}\left( {{e_{n + m}}(k)} \right)} \right)$,\\
which can be further written as (see the Appendix \ref{sec:a1} for a detailed derivation):
\begin{equation}\label{equ:33}
{\bf{x}}(k) = \widehat {\bf{x}}(k|k - 1) + \overline {\bf{K}} (k)\left( {{\bf{y}}(k) - \widehat {\bf{y}}(k)} \right)
\end{equation}
where
\begin{equation}\label{equ:34}
\resizebox{1\hsize}{!}
{$\begin{cases}
\overline {\bf{K}} (k) = \overline {\bf{P}} (k|k - 1){{\bf{H}}^T}(k){\left( {{\bf{H}}(k)\overline {\bf{P}} (k|k - 1){{\bf{H}}^T}(k) + \overline {\bf{R}} (k)} \right)^{ - 1}}\\
\overline {\bf{P}} (k|k - 1) = {{\bf{S}}_p}(k|k - 1){\bf{C}}_x^{ - 1}(k){\bf{S}}_p^T(k|k - 1)\\
\overline {\bf{R}} (k) = {{\bf{S}}_r}(k){\bf{C}}_y^{ - 1}(k){\bf{S}}_r^T(k)
\end{cases}$}
\end{equation}
Meanwhile, the corresponding covariance matrix is updated by
\begin{equation}\label{equ:35}
\resizebox{0.89\hsize}{!}
{$\begin{split}
{\bf{P}}(k|k) = \ &\left( {{\bf{I}} - \overline {\bf{K}} (k){\bf{H}}(k)} \right){\bf{P}}(k|k - 1){\left( {{\bf{I}} - \overline {\bf{K}} (k){\bf{H}}(k)} \right)^T}\\
&+\overline {\bf{K}} (k){\bf{R}}(k){\overline {\bf{K}} ^T}(k)
\end{split}$}
\end{equation}
\begin{remark}\label{rem:1}
Since $\overline {\bf{K}} (k)$ relies on $\overline {\bf{P}} (k|k - 1)$ and $\overline {\bf{R}} (k)$, both related to ${\bf{x}}(k)$ through ${{\bf{C}}_x}(k)$ and ${{\bf{C}}_y}(k)$, respectively, the equation (\ref{equ:33}) is a fixed-point equation of ${\bf{x}}(k)$. One can solve (\ref{equ:33}) by a fixed-point iterative method. The initial value of the fixed-point iteration can be set to $\widehat {\bf{x}}{(k|k)_0} = \widehat {\bf{x}}(k|k - 1)$ or chosen as the least-squares solution $\widehat {\bf{x}}{(k|k)_0} = {\left( {{{\bf{W}}^T}(k){\bf{W}}(k)} \right)^{ - 1}}{{\bf{W}}^T}(k){\bf{D}}(k)$. The value after convergence is the posterior estimate of the state $\widehat {\bf{x}}(k|k)$.
\end{remark}

A detailed description of the proposed MCUF algorithm is as follows:
\begin{enumerate}
\item[1)] Choose a proper kernel bandwidth $\sigma $ and a small positive $\varepsilon$; Set an initial estimate $\widehat {\bf{x}}(0|0)$ and corresponding covariance matrix ${\bf{P}}(0|0)$; Let $k = 1$;
\item[2)] Use equations (\ref{equ:9})$\sim$(\ref{equ:14}) to obtain prior estimate $\widehat {\bf{x}}(k|k - 1)$ and covariance ${\bf{P}}(k|k - 1)$, and calculate ${{\bf{S}}_p}(k|k - 1)$ by Cholesky decomposition ;
\item[3)] Use (\ref{equ:14})$\sim$(\ref{equ:17}) to compute the prior measurement $\widehat {\bf{y}}(k)$ and use (\ref{equ:13}) (\ref{equ:18}) and (\ref{equ:20}) to acquire the measurement slope matrix ${\bf{H}}(k)$, and construct the statistical linear regression model (\ref{equ:22});
\item[4)] Transform (\ref{equ:22}) into (\ref{equ:24}), and let $t=1$ and $\widehat {\bf{x}}{(k|k)_0} = {\left( {{{\bf{W}}^T}(k){\bf{W}}(k)} \right)^{ - 1}}{{\bf{W}}^T}(k){\bf{D}}(k)$;
\item[5)] Use (\ref{equ:36})$\sim$(\ref{equ:42}) to compute $\widehat {\bf{x}}{(k|k)_t}$;
 \begin{equation}\label{equ:36}
\widehat {\mathbf{x}}{(k|k)_t} = \widehat {\mathbf{x}}(k|k - 1) + \widetilde {\mathbf{K}}(k)\left( {{\mathbf{y}}(k) - {\widehat {\bf{y}}(k)} } \right)
\end{equation}
with
\begin{equation}\label{equ:37}
\resizebox{0.90\hsize}{!}
{$\widetilde {\mathbf{K}}(k) = \widetilde {\mathbf{P}}(k|k - 1){{\mathbf{H}}^T}(k){\left( {{\mathbf{H}}(k)\widetilde {\mathbf{P}}(k|k - 1){{\mathbf{H}}^T}(k) + \widetilde {\mathbf{R}}(k)} \right)^{ - 1}},$}
\end{equation}
\begin{equation}\label{equ:38}
\widetilde {\mathbf{P}}(k|k - 1) = {{\mathbf{S}}_p}(k|k - 1)\widetilde {\mathbf{C}}_x^{ - 1}(k){\mathbf{S}}_p^T(k|k - 1),
\end{equation}
\begin{equation}\label{equ:39}
\widetilde {\mathbf{R}}(k) = {{\mathbf{S}}_r}(k)\widetilde {\mathbf{C}}_y^{ - 1}(k){\mathbf{S}}_r^T(k).
\end{equation}
\begin{equation}\label{equ:40}
{\widetilde {\mathbf{C}}_x}(k) = diag\left( {{{\operatorname G}_\sigma }\left( {{{\widetilde e}_1}(k)} \right),...,{{\operatorname G}_\sigma }\left( {{{\widetilde e}_n}(k)} \right)} \right)
\end{equation}
\begin{equation}\label{equ:41}
{\widetilde {\mathbf{C}}_y}(k) = diag\left( {{{\operatorname G}_\sigma }\left( {{{\widetilde e}_{n + 1}}(k)} \right),...,{{\operatorname G}_\sigma}\left( {{{\widetilde e}_{n + m}}(k)} \right)} \right)
\end{equation}
\begin{equation}\label{equ:42}
{\widetilde e_i}(k) = {d_i}(k) - {{\mathbf{w}}_i}(k)\widehat {\mathbf{x}}{(k|k)_{t - 1}}
\end{equation}
\item[6)] Compare the estimation at the current step and the estimation at the last step. If (\ref{equ:43}) holds, set $\widehat {\mathbf{x}}(k|k) = \widehat {\mathbf{x}}{(k|k)_t}$ and continue to step 7); Otherwise, $t + 1 \to t$, and go back to step 5).
\begin{equation}\label{equ:43}
\frac{{\left\| {\widehat {\mathbf{x}}{{(k|k)}_t} - \widehat {\mathbf{x}}{{(k|k)}_{t - 1}}} \right\|}}{{\left\| {\widehat {\mathbf{x}}{{(k|k)}_{t - 1}}} \right\|}} \le \varepsilon
\end{equation}
\item[7)] Update the posterior covariance matrix by (\ref{equ:44}), $k + 1 \to k$ and go back to step 2).
\begin{equation}\label{equ:44}
\resizebox{0.90\hsize}{!}
{$\begin{split}
{\mathbf{P}}(k|k) = &\left( {{\mathbf{I}} - \widetilde {\mathbf{K}}(k){\mathbf{H}}(k)} \right){\mathbf{P}}(k|k - 1){\left( {{\mathbf{I}} - \widetilde {\mathbf{K}}(k){\bf{H}}(k)} \right)^T}\\
 &+\widetilde {\bf{K}}(k){\bf{R}}(k){\widetilde {\bf{K}}^T}(k)
\end{split}$}
\end{equation}
\end{enumerate}
\begin{remark}\label{rem:2}
As one can see, (\ref{equ:9})$\sim$(\ref{equ:18}) are the unscented transformation (UT). In the MCUF, we use a statistical linear regression model and the MCC to obtain the posterior estimates of the state and covariance. With the UT and statistical linear approximation, the proposed filter can achieve a more accurate solution than the first order linearization based filters. Moreover, the usage of MCC will improve the robustness of the filter against large outliers. Usually the convergence of the fixed-point iteration to the optimal solution is very fast (see Section \ref{sec:4}). Thus, the computational complexity of MCUF is not high. The kernel bandwidth $\sigma $, which can be set manually or optimized by trial and error methods in practical applications, is a key parameter in MCUF. In general, a smaller kernel bandwidth makes the algorithm more robust (with respect to outliers), but a too small kernel bandwidth may lead to slow convergence or even divergence of the algorithm. From \cite{Chen2015convergence}, we know that if the kernel bandwidth is larger than a certain value, the fixed-point equation (\ref{equ:29}) will surely converge to a unique fixed point. When $\sigma  \to \infty $, the MCUF will obtain the least-squares solution ${\left( {{{\bf{W}}^T}(k){\bf{W}}(k)} \right)^{ - 1}}{{\bf{W}}^T}(k){\bf{D}}(k)$.
\end{remark}

\section{Illustrative examples}
\label{sec:4}

In this section, we present two illustrative examples to demonstrate the performance of the proposed MCUF algorithm. Moreover, the performance is measured using the following benchmarks:
\begin{equation}\label{equ:45}
{{\mathop{\rm MSE}} _1}(k) = \frac{1}{M}\sum\limits_{m = 1}^M {{{(x(k) - \widehat x(k|k))}^2}} ,\ {\text{for}}\ k = 1 \ldots K
\end{equation}
\begin{equation}\label{equ:46}
{{\mathop{\rm MSE}} _2}(m) = \frac{1}{K}\sum\limits_{k = 1}^K {{{(x(k) - \widehat x(k|k))}^2}} ,\ {\text{for}}\ m = 1 \ldots M
\end{equation}
\begin{equation}\label{equ:47}
{\mathop{\rm MSE}} = \frac{1}{M}\sum\limits_{m = 1}^M {{{{\mathop{\rm MSE}} }_2}(m)} = \frac{1}{K}\sum\limits_{k = 1}^K {{{{\mathop{\rm MSE}} }_1}(k)}
\end{equation}
where $K$ is the total time steps in every Monte Carlo run and $M$ represents the total number of Monte Carlo runs.

\subsection{Example 1}
\label{subsec:41}

Consider the univariate nonstationary growth model (UNGM), which is often used as a benchmark example for nonlinear filtering. The state and measurement equations are given by
\begin{equation}\label{equ:48}
\begin{split}
x(k) = \ &0.5x(k - 1) + 25\frac{{x(k - 1)}}{{1 + x{{(k - 1)}^2}}}\\
&+ 8\cos \left( {1.2(k - 1)} \right) + q(k - 1),
\end{split}
\end{equation}
\begin{equation}\label{equ:49}
y(k) = \frac{{x{{(k)}^2}}}{{20}} + r(k).
\end{equation}
First, we consider the case in which the noises are all Gaussian, that is,
\begin{equation*}
\begin{array}{l}
q(k - 1) \sim N(0,1)\\
r(k) \sim N(0,1)
\end{array}
\end{equation*}

Table \ref{tab:1} lists the ${\mathop{\rm MSE}}$s of $x$, defined in (\ref{equ:47}), and the average fixed-point iteration numbers. In the simulation, the parameters are set as $K=500, M=100$. Since all the noises are Gaussian, the UKF achieves the smallest MSE among all the filters. In this example, we should choose a larger kernel bandwidth in MCUF to have a good performance. One can also observe that the average iteration numbers of the MCUF are relatively small especially when the kernel bandwidth is large.
\begin{table}[htbp]
\renewcommand{\arraystretch}{1.3}
\caption{${\mathop{\rm MSE}}$s of $x$ and Average Iteration Numbers in Gaussian Noises}
\label{tab:1}
\centering
\begin{tabular}{ccc}
\hline
Filter& ${\mathop{\rm MSE}}$ of $x$ & Average iteration number\\
\hline
UKF & 68.9766 & ---\\
MCUF$\left( {\sigma  = 2.0,\varepsilon  = {{10}^{ - 6}}} \right)$ & 108.9796 & 5.0624\\
MCUF$\left( {\sigma  = 3.0,\varepsilon  = {{10}^{ - 6}}} \right)$ & 94.5856 & 4.5431 \\
MCUF$\left( {\sigma  = 5.0,\varepsilon  = {{10}^{ - 6}}} \right)$ & 83.7554 & 3.7323 \\
MCUF$\left( {\sigma  = 8.0,\varepsilon  = {{10}^{ - 6}}} \right)$ & 86.1612 & 3.0431\\
MCUF$\left( {\sigma  = 10,\varepsilon  = {{10}^{ - 6}}} \right)$ & 85.4109 & 2.7919\\
\hline
\end{tabular}
\end{table}

Second, we consider the case in which the process noise is still Gaussian but the measurement noise is a heavy-tailed (impulsive) non-Gaussian noise, with a mixed-Gaussian distribution, that is,
\begin{equation*}
\begin{array}{l}
q(k - 1) \sim N(0,1)\\
r(k) \sim 0.8N(0,1) + 0.2N(0,400)
\end{array}
\end{equation*}

Table \ref{tab:2} illustrates the corresponding ${\mathop{\rm MSE}}$s of $x$ and average iteration numbers. As one can see, in impulsive noises, when kernel bandwidth is too small or too large, the performance of MCUF will be not good. However, with a proper kernel bandwidth (say $\sigma=2.0$), the MCUF can outperform all the filters, achieving the smallest MSE. In addition, it is evident that the larger the kernel bandwidth, the faster the convergence speed. In general, the fixed-point algorithm in MCUF will converge to the optimal solution in only few iterations.

\begin{table}[htbp]
\renewcommand{\arraystretch}{1.3}
\caption{${\mathop{\rm MSE}}$s of $x$ and Average Iteration Numbers in Gaussian Process Noise and Non-Gaussian Measurement Noise}
\label{tab:2}
\centering
\begin{tabular}{ccc}
\hline
Filter& ${\mathop{\rm MSE}}$ of $x$ & Average iteration number\\
\hline
UKF & 84.3496 & ---\\
MCUF$\left( {\sigma  = 1.0,\varepsilon  = {{10}^{ - 6}}} \right)$ & 70.5870 & 3.5413\\
MCUF$\left( {\sigma  = 2.0,\varepsilon  = {{10}^{ - 6}}} \right)$ & 68.9714 & 3.0352 \\
MCUF$\left( {\sigma  = 3.0,\varepsilon  = {{10}^{ - 6}}} \right)$ & 69.3548 & 2.7056 \\
MCUF$\left( {\sigma  = 5.0,\varepsilon  = {{10}^{ - 6}}} \right)$ & 69.4932 & 2.3391\\
MCUF$\left( {\sigma  = 10,\varepsilon  = {{10}^{ - 6}}} \right)$ & 70.4666 & 2.0286\\
\hline
\end{tabular}
\end{table}

We also investigate the influence of the threshold $\varepsilon$ on the performance. The ${\mathop{\rm MSE}}$s of $x$ and average iteration numbers with different $\varepsilon$ (The kernel bandwidth is set at $\sigma=2.0$) are given in Table \ref{tab:3}. Usually, a smaller $\varepsilon$ results in a slightly lower MSE but a larger iteration number for convergence. Without mentioned otherwise, we choose ${\varepsilon  = {{10}^{ - 6}}}$ in this work.

\begin{table}[htbp]
\renewcommand{\arraystretch}{1.3}
\caption{${\mathop{\rm MSE}}$s of $x$ and Average Iteration Numbers with Different $\varepsilon$}
\label{tab:3}
\centering
\begin{tabular}{ccc}
\hline
Filter& ${\mathop{\rm MSE}}$ of $x$ & Average iteration number\\
\hline
MCUF$\left( {\sigma  = 2.0,\varepsilon  = {{10}^{ - 1}}} \right)$ & 69.7465 & 1.1142\\
MCUF$\left( {\sigma  = 2.0,\varepsilon  = {{10}^{ - 2}}} \right)$ & 69.5853 & 1.3777\\
MCUF$\left( {\sigma  = 2.0,\varepsilon  = {{10}^{ - 4}}} \right)$ & 69.0197 & 2.1753\\
MCUF$\left( {\sigma  = 2.0,\varepsilon  = {{10}^{ - 6}}} \right)$ & 68.8073 & 3.0282\\
MCUF$\left( {\sigma  = 2.0,\varepsilon  = {{10}^{ - 8}}} \right)$ & 68.8148 & 3.8770\\
\hline
\end{tabular}
\end{table}

Further, we consider the situation where the process and measurement noises are all non-Gaussian with mixed-Gaussian distributions:
\begin{equation*}
\begin{array}{l}
q(k - 1) \sim 0.8N(0,0.1) + 0.2N(0,10)\\
r(k) \sim 0.8N(0,1) + 0.2N(0,400)
\end{array}
\end{equation*}

With the same parameters setting as before, the results are presented in Table \ref{tab:4}. As expected, with a proper kernel bandwidth the MCUF can achieve the best performance.

\begin{table}[htbp]
\renewcommand{\arraystretch}{1.3}
\caption{${\mathop{\rm MSE}}$s of $x$ and Average Iteration Numbers in Non-Gaussian Process Noise and Measurement Noise}
\label{tab:4}
\centering
\begin{tabular}{ccc}
\hline
Filter& ${\mathop{\rm MSE}}$ of $x$ & Average iteration number\\
\hline
UKF & 84.8735 & ---\\
MCUF$\left( {\sigma  = 1.0,\varepsilon  = {{10}^{ - 6}}} \right)$ & 71.7599 & 3.6104\\
MCUF$\left( {\sigma  = 2.0,\varepsilon  = {{10}^{ - 6}}} \right)$ & 69.4382 & 3.1142 \\
MCUF$\left( {\sigma  = 3.0,\varepsilon  = {{10}^{ - 6}}} \right)$ & 69.8497 & 2.7765 \\
MCUF$\left( {\sigma  = 5.0,\varepsilon  = {{10}^{ - 6}}} \right)$ & 69.8505 & 2.3879\\
MCUF$\left( {\sigma  = 10,\varepsilon  = {{10}^{ - 6}}} \right)$ & 70.0356 & 2.0599\\
\hline
\end{tabular}
\end{table}

\subsection{Example 2}
\label{subsec:42}

In this example, we consider a practical model \cite{Simon2006optimal}. and the performance of EKF \cite{Anderson1979optimal}, Huber-EKF (HEKF) \cite{Hawary1995robust}, UKF \cite{Julier2000a} and HUKF \cite{Wang2010huber} are also presented for comparison purpose. The goal is to estimate the position, velocity and ballistic coefficient of a vertically falling body at a very high altitude. The measurements are taken by a radar system each $0.1s$. By rectangle integral in the discrete time with sampling period $\Delta T$, we obtain the following model:
\begin{equation}\label{equ:50}
\begin{split}
{x_1}({k_1}) = \ &{x_1}({k_1} - 1) + \Delta T{x_2}({k_1} - 1) + {q_1}({k_1} - 1)\\
{x_2}({k_1}) = \ &{x_2}({k_1} - 1) + \Delta T{\rho _0}\exp \left( { - {x_1}({k_1} - 1)/a} \right)\\
\times&{x_2}{({k_1} - 1)^2}{x_3}({k_1} - 1)/2 - \Delta Tg + {q_2}({k_1} - 1)\\
{x_3}({k_1}) = \ &{x_3}({k_1} - 1) + {q_3}({k_1} - 1)
\end{split}
\end{equation}
\begin{equation}\label{equ:51}
y(k) = \sqrt {{b^2} + {{({x_1}(k) - H)}^2}}  + r(k)
\end{equation}
where the sampling time is $\Delta T = 0.001s$, that is $k=100k_1$, the constant ${\rho _0}$ is ${\rho _0} = 2$, the constant $a$ that relates the air density with altitude is $a = 20000$, the acceleration of gravity is $g = 32.2ft/s^2$, the located altitude of radar is $H = 100000ft$, and the horizontal range between the body and the radar is $b = 100000ft$.

The state vector ${\mathbf{x}}(k) = {\left[ {\begin{array}{*{20}{c}}
{{x_1}(k)}&{{x_2}(k)}&{{x_3}(k)}
\end{array}} \right]^T}$ contains the position, velocity and ballistic coefficient. Similar to \cite{Simon2006optimal}, we do not introduce any process noise in this model. The initial state is assumed to be ${\mathbf{x}}(0) = {\left[ {\begin{array}{*{20}{c}}
300000&-20000&1/1000\end{array}} \right]^T}$, and the initial estimate is $\widehat {\mathbf{x}}(0|0) = {\left[ {\begin{array}{*{20}{c}}
300000&-20000&0.0009\end{array}} \right]^T}$ with covariance matrix ${\mathbf{P}}(0|0) = diag([1000000,4000000,1/1000000])$.

First, we assume that the measurement noise is Gaussian, that is,
\begin{equation*}
\begin{array}{l}
{q_1}({k_1}) \sim 0\\
{q_2}({k_1}) \sim 0\\
{q_3}({k_1}) \sim 0\\
r(k) \sim N(0,10000)
\end{array}
\end{equation*}

\begin{figure}[htbp]
\centering
\includegraphics[width=3in]{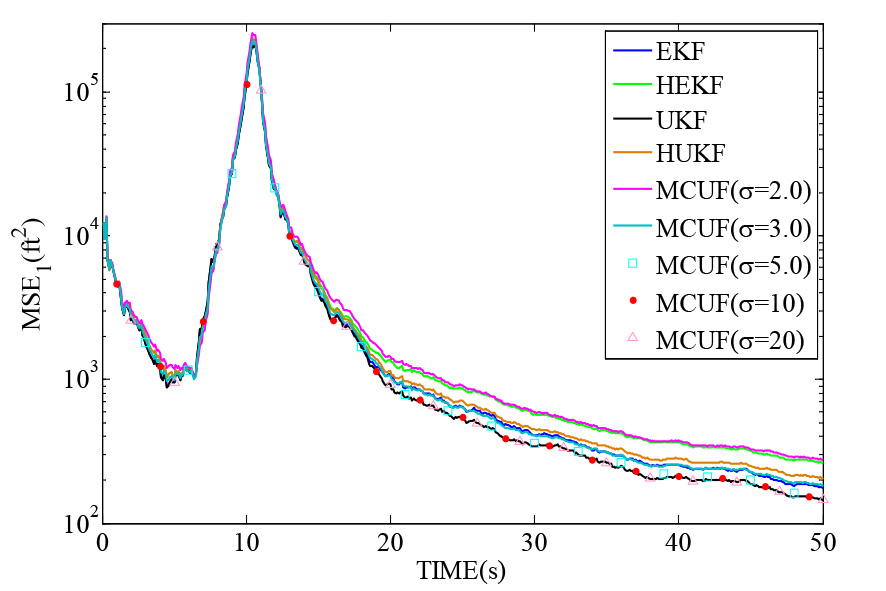}
\centering
\caption{${\mathop{\rm MSE}}_{1}$ of $x_1$ in Gaussian noise}
\label{fig:1}
\end{figure}

\begin{figure}[htbp]
\centering
\includegraphics[width=3in]{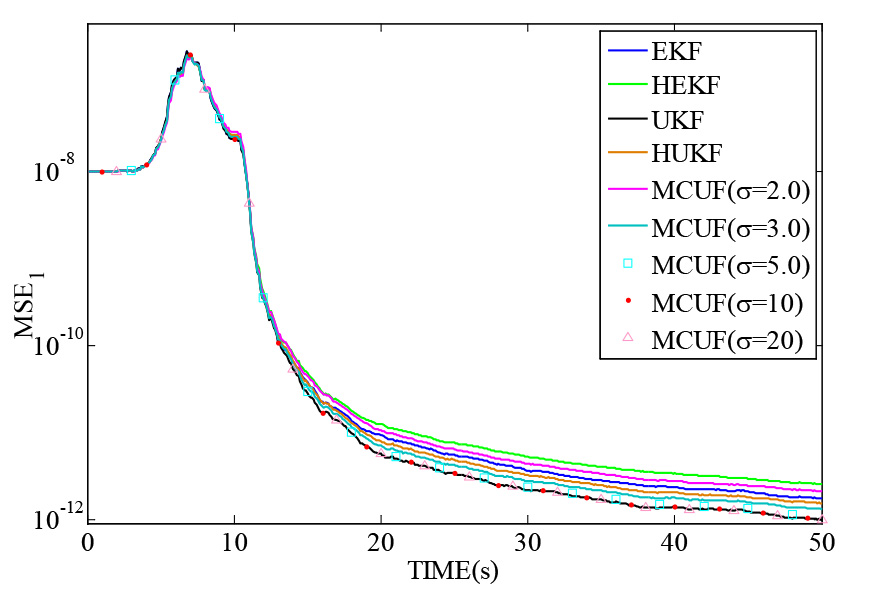}
\centering
\caption{${\mathop{\rm MSE}}_{1}$ of $x_3$ in Gaussian noise}
\label{fig:3}
\end{figure}

In the simulation, we consider the motion during the first $50s$, and make $100$ independent Monte Carlo runs, that is, $K=500, M=100$. Fig. \ref{fig:1} $\sim$ Fig. \ref{fig:3} show the ${{\mathop{\rm MSE}} _1}$ (as defined in (\ref{equ:45})) of $x_1$, $x_2$ and $x_3$ for different filters in Gaussian noise. The corresponding ${\mathop{\rm MSE}}$s and average fixed-point iteration numbers are summarized in Table \ref{tab:5} and Table \ref{tab:6}. As one can see, in this case, since the noise is Gaussian, the UKF performs the best. However, the proposed MCUF with large kernel bandwidth can outperform the other two robust Kalman type filters, namely HEKF and HUKF. One can also observe that the average iteration numbers of the MCUF are very small especially when the kernel bandwidth is large.

\begin{figure}[htbp]
\centering
\includegraphics[width=3in]{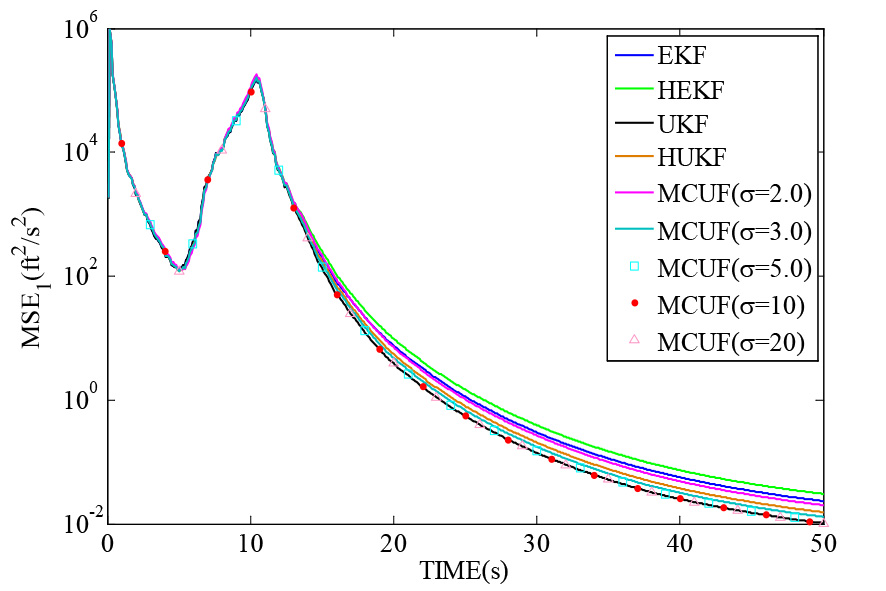}
\centering
\caption{${\mathop{\rm MSE}}_{1}$ of $x_2$ in Gaussian noise}
\label{fig:2}
\end{figure}
\begin{table}[htbp]
\renewcommand{\arraystretch}{1.3}
\caption{${\mathop{\rm MSE}}$s of $x_1$, $x_2$ and $x_3$ in Gaussian Noise}
\label{tab:5}
\centering
\resizebox{1.0\hsize}{!}
{$\begin{tabular}{cccc}
\hline
Filter& ${\mathop{\rm MSE}}$ of $x_1$ & ${\mathop{\rm MSE}}$ of $x_2$ & ${\mathop{\rm MSE}}$ of $x_3$\\
\hline
EKF & $7.3254 \times {10^3}$ & $8.6071 \times {10^3}$ & $1.1317 \times {10^{-8}}$\\
HEKF$\left( {\varepsilon  = {{10}^{ - 6}}} \right)$ & $7.9266 \times {10^3}$ & $8.9799 \times {10^3}$ & $1.1001 \times {10^{-8}}$\\
UKF & $7.2630 \times {10^3}$ & $8.5948 \times {10^3}$ & $1.1308 \times {10^{-8}}$\\
HUKF$\left( {\varepsilon  = {{10}^{ - 6}}} \right)$ & $7.8343 \times {10^3}$ & $8.9969 \times {10^3}$ & $1.0983 \times {10^{-8}}$\\
MCUF$\left( {\sigma  = 2.0,\varepsilon  = {{10}^{ - 6}}} \right)$ & $8.3984 \times {10^3}$ & $9.4045 \times {10^3}$ & $1.0864 \times {10^{-8}}$\\
MCUF$\left( {\sigma  = 3.0,\varepsilon  = {{10}^{ - 6}}} \right)$ & $7.4680 \times {10^3}$ & $8.7892 \times {10^3}$ & $1.0865 \times {10^{-8}}$\\
MCUF$\left( {\sigma  = 5.0,\varepsilon  = {{10}^{ - 6}}} \right)$ & $7.2943 \times {10^3}$ & $8.6564 \times {10^3}$ & $1.1098 \times {10^{-8}}$\\
MCUF$\left( {\sigma  = 10,\varepsilon  = {{10}^{ - 6}}} \right)$ & $7.2674 \times {10^3}$ & $8.6298 \times {10^3}$ & $1.1246 \times {10^{-8}}$\\
MCUF$\left( {\sigma  = 20,\varepsilon  = {{10}^{ - 6}}} \right)$ & $7.2642 \times {10^3}$ & $8.6254 \times {10^3}$ & $1.1288 \times {10^{-8}}$\\
\hline
\end{tabular}$}
\end{table}
\begin{table}[htbp]
\renewcommand{\arraystretch}{1.3}
\caption{Average Iteration Numbers for Every Time Step in Gaussian Noise}
\label{tab:6}
\centering
\begin{tabular}{cc}
\hline
Filter& Average iteration number\\
\hline
HEKF$\left( {\varepsilon  = {{10}^{ - 6}}} \right)$ & 1.2342\\
HUKF$\left( {\varepsilon  = {{10}^{ - 6}}} \right)$ & 1.2332\\
MCUF$\left( {\sigma  = 2.0,\varepsilon  = {{10}^{ - 6}}} \right)$ & 1.7881\\
MCUF$\left( {\sigma  = 3.0,\varepsilon  = {{10}^{ - 6}}} \right)$ & 1.5821\\
MCUF$\left( {\sigma  = 5.0,\varepsilon  = {{10}^{ - 6}}} \right)$ & 1.3808\\
MCUF$\left( {\sigma  = 10,\varepsilon  = {{10}^{ - 6}}} \right)$ & 1.1706\\
MCUF$\left( {\sigma  = 20,\varepsilon  = {{10}^{ - 6}}} \right)$ & 1.0521\\
\hline
\end{tabular}
\end{table}

Second, we consider the case in which the measurement noise is a heavy-tailed (impulsive) non-Gaussian noise, with a mixed-Gaussian distribution, that is,
\begin{equation*}
\begin{array}{l}
{q_1}({k_1}) \sim 0\\
{q_2}({k_1}) \sim 0\\
{q_3}({k_1}) \sim 0\\
r(k) \sim 0.7N(0,1000) + 0.3N(0,100000)
\end{array}
\end{equation*}

\begin{figure}[htbp]
\centering
\includegraphics[width=3in]{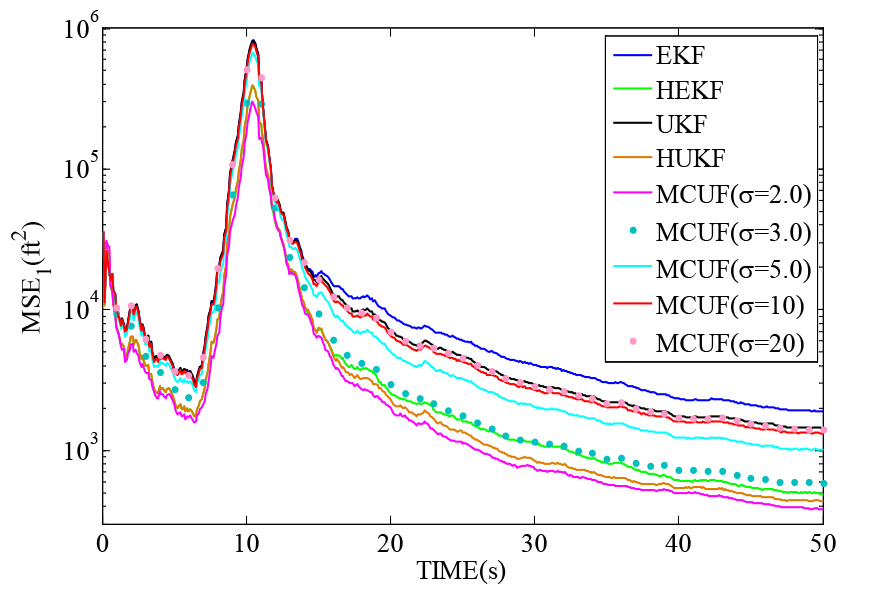}
\centering
\caption{${\mathop{\rm MSE}}_{1}$ of $x_1$ in non-Gaussian noise}
\label{fig:4}
\end{figure}

\begin{figure}[htbp]
\centering
\includegraphics[width=3in]{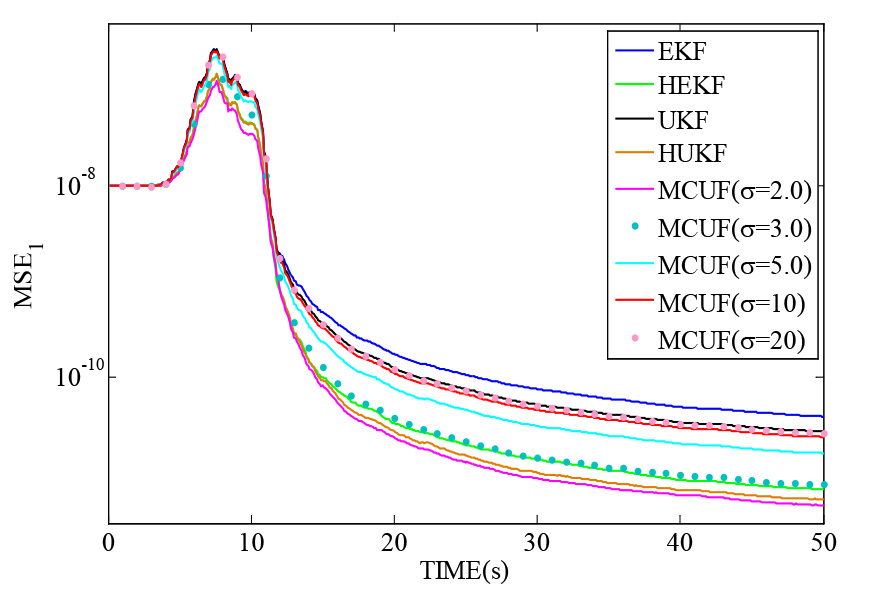}
\centering
\caption{${\mathop{\rm MSE}}_{1}$ of $x_3$ in non-Gaussian noise}
\label{fig:6}
\end{figure}

Fig. \ref{fig:4} $\sim$ Fig. \ref{fig:6} demonstrate the ${\mathop{\rm MSE}}_{1}$ of $x_1$, $x_2$ and $x_3$ for different filters in non-Gaussian noise, and Table \ref{tab:7} and Table \ref{tab:8} summarize the corresponding ${\mathop{\rm MSE}}$s and average iteration numbers respectively. We can see clearly that the three robust Kalman type filters (HEKF, HUKF, MCUF) are superior to their non-robust counterparts (EKF, UKF). When the kernel bandwidth is very large, the MCUF achieves almost the same performance as that of UKF. In contrast, with a smaller kernel bandwidth, the MCUF can outperform the UKF significantly. Especially, when $\sigma  = 2.0$, the MCUF exhibits the smallest ${\mathop{\rm MSE}}$ among all the algorithms. Again, the fixed-point algorithm in MCUF will converge to the optimal solution in very few iterations.

\begin{figure}[htbp]
\centering
\includegraphics[width=3in]{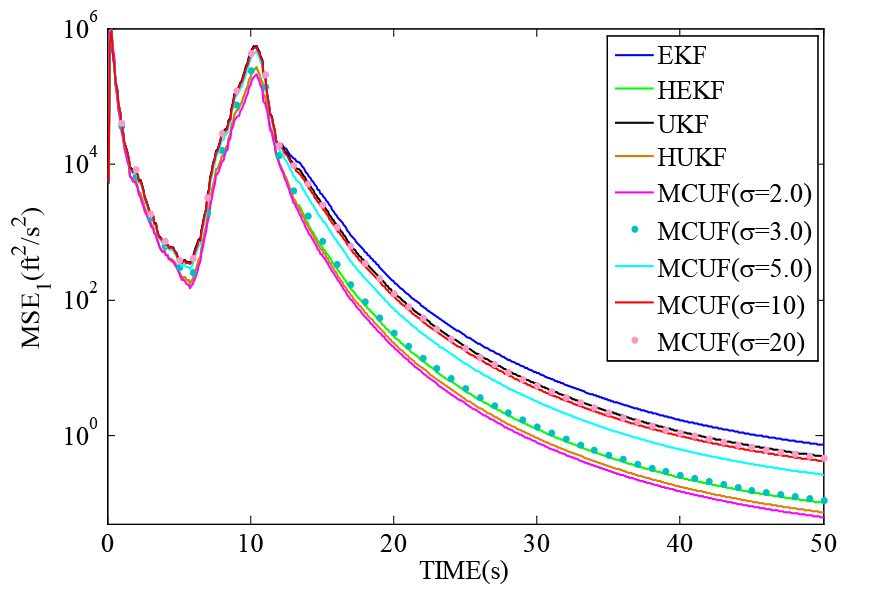}
\centering
\caption{${\mathop{\rm MSE}}_{1}$ of $x_2$ in non-Gaussian noise}
\label{fig:5}
\end{figure}
\begin{table}[htbp]
\renewcommand{\arraystretch}{1.3}
\caption{${\mathop{\rm MSE}}$s of $x_1$, $x_2$ and $x_3$ in Non-Gaussian Noise}
\label{tab:7}
\centering
\resizebox{1.0\hsize}{!}
{$\begin{tabular}{cccc}
\hline
Filter& ${\mathop{\rm MSE}}$ of $x_1$ & ${\mathop{\rm MSE}}$ of $x_2$ & ${\mathop{\rm MSE}}$ of $x_3$\\
\hline
EKF & $2.9499 \times {10^4}$ & $2.2283 \times {10^4}$ & $1.5566 \times {10^{-8}}$\\
HEKF$\left( {\varepsilon  = {{10}^{ - 6}}} \right)$ & $1.4068 \times {10^4}$ & $1.4079 \times {10^4}$ & $8.6645 \times {10^{-9}}$\\
UKF & $2.8772 \times {10^4}$ & $2.2331 \times {10^4}$ & $1.5497 \times {10^{-8}}$\\
HUKF$\left( {\varepsilon  = {{10}^{ - 6}}} \right)$ & $1.3996 \times {10^4}$ & $1.4212 \times {10^4}$ & $8.6247 \times {10^{-9}}$\\
MCUF$\left( {\sigma  = 2.0,\varepsilon  = {{10}^{ - 6}}} \right)$ & $1.1457 \times {10^4}$ & $1.2990 \times {10^4}$ & $7.3965 \times {10^{-9}}$\\
MCUF$\left( {\sigma  = 3.0,\varepsilon  = {{10}^{ - 6}}} \right)$ & $1.7612 \times {10^4}$ & $1.5970 \times {10^4}$ & $1.0199 \times {10^{-8}}$\\
MCUF$\left( {\sigma  = 5.0,\varepsilon  = {{10}^{ - 6}}} \right)$ & $2.3818 \times {10^4}$ & $1.9475 \times {10^4}$ & $1.3083 \times {10^{-8}}$\\
MCUF$\left( {\sigma  = 10,\varepsilon  = {{10}^{ - 6}}} \right)$ & $2.7448 \times {10^4}$ & $2.1572 \times {10^4}$ & $1.4826 \times {10^{-8}}$\\
MCUF$\left( {\sigma  = 20,\varepsilon  = {{10}^{ - 6}}} \right)$ & $2.8428 \times {10^4}$ & $2.2147 \times {10^4}$ & $1.5323 \times {10^{-8}}$\\
\hline
\end{tabular}$}
\end{table}
\begin{table}[htbp]
\renewcommand{\arraystretch}{1.3}
\caption{Average Iteration Numbers for Every Time Step in Non-Gaussian Noise}
\label{tab:8}
\centering
\begin{tabular}{cc}
\hline
Filter& Average iteration number\\
\hline
HEKF$\left( {\varepsilon  = {{10}^{ - 6}}} \right)$ & 1.2280\\
HUKF$\left( {\varepsilon  = {{10}^{ - 6}}} \right)$ & 1.2300\\
MCUF$\left( {\sigma  = 2.0,\varepsilon  = {{10}^{ - 6}}} \right)$ & 1.4809\\
MCUF$\left( {\sigma  = 3.0,\varepsilon  = {{10}^{ - 6}}} \right)$ & 1.3815\\
MCUF$\left( {\sigma  = 5.0,\varepsilon  = {{10}^{ - 6}}} \right)$ & 1.2771\\
MCUF$\left( {\sigma  = 10,\varepsilon  = {{10}^{ - 6}}} \right)$ & 1.1713\\
MCUF$\left( {\sigma  = 20,\varepsilon  = {{10}^{ - 6}}} \right)$ & 1.0917\\
\hline
\end{tabular}
\end{table}

\section{Conclusion}
\label{sec:5}
In this work, we propose a novel nonlinear Kalman type filter, namely the maximum correntropy unscented filter (MCUF), by using the unscented transformation (UT) to get the prior estimates of the state and covariance matrix and applying a statistical linearization regression model based on the maximum correntropy criterion (MCC) to obtain the posterior estimates (solved by a fixed-point iteration) of the state and covariance. Simulation results demonstrate that with a  proper kernel bandwidth, the MCUF can achieve better performance than some existing algorithms including EKF, HEKF, UKF and HUKF particularly when the underlying system is disturbed by some impulsive noises.


%

\appendices
\section{Derivation of (\ref{equ:33})}
\label{sec:a1}
\begin{equation}\label{equ:a1}
\begin{split}
{\mathbf{W}}\left( k \right) & = {{\mathbf{S}}^{ - 1}}\left( k \right)\left[ {\begin{array}{*{20}{c}}
{\mathbf{I}}\\
{{\mathbf{H}}\left( k \right)}
\end{array}} \right]\\
& = \left[ {\begin{array}{*{20}{c}}
{{\mathbf{S}}_p^{ - 1}\left( {k|k - 1} \right)}&0\\
0&{{\mathbf{S}}_r^{ - 1}\left( k \right)}
\end{array}} \right]\left[ {\begin{array}{*{20}{c}}
{\mathbf{I}}\\
{{\mathbf{H}}\left( k \right)}
\end{array}} \right]\\
& = \left[ {\begin{array}{*{20}{c}}
{{\mathbf{S}}_p^{ - 1}\left( {k|k - 1} \right)}\\
{{\mathbf{S}}_r^{ - 1}\left( k \right){\mathbf{H}}\left( k \right)}
\end{array}} \right]
\end{split}
\end{equation}
\begin{equation}\label{equ:a2}
{\mathbf{C}}\left( k \right) = \left[ {\begin{array}{*{20}{c}}
  {{{\mathbf{C}}_x}\left( k \right)}&0 \\
  0&{{{\mathbf{C}}_y}\left( k \right)}
\end{array}} \right]
\end{equation}
\begin{equation}\label{equ:a3}
\resizebox{1.0\hsize}{!}
{$\begin{split}
  {\mathbf{D}}\left( k \right)& = {{\mathbf{S}}^{ - 1}}\left( k \right)\left[ {\begin{array}{*{20}{c}}
  {\widehat {\mathbf{x}}\left( {k|k - 1} \right)} \\
  {{\mathbf{y}}\left( k \right){ - {\mathop{\operatorname h}\nolimits} \left( {k,\widehat {\mathbf{x}}(k|k - 1)} \right) + {\mathbf{H}}(k)\widehat {\mathbf{x}}(k|k - 1)}}
\end{array}} \right] \\
  & = \left[ {\begin{array}{*{20}{c}}
  {{\mathbf{S}}_p^{ - 1}\left( {k|k - 1} \right)\widehat {\mathbf{x}}\left( {k|k - 1} \right)} \\
  {{\mathbf{S}}_r^{ - 1}{\left( {{\mathbf{y}}(k) - {\mathop{\operatorname h}\nolimits} \left( {k,\widehat {\mathbf{x}}(k|k - 1)} \right) + {\mathbf{H}}(k)\widehat {\mathbf{x}}(k|k - 1)} \right)}}
\end{array}} \right] \\
\end{split}$}
\end{equation}
By (\ref{equ:a1}) and (\ref{equ:a2}), we have
\begin{equation}\label{equ:a4}
\begin{split}
  &{\left( {{{\mathbf{W}}^T}\left( k \right){\mathbf{C}}\left( k \right){\mathbf{W}}\left( k \right)} \right)^{ - 1}} \hfill \\
   = & {\left[ {{{\left( {{\mathbf{S}}_p^{ - 1}} \right)}^T}{{\mathbf{C}}_x}{\mathbf{S}}_p^{ - 1} + {{\mathbf{H}}^T}{{\left( {{{\mathbf{S}}_r}^{ - 1}} \right)}^T}{{\mathbf{C}}_y}{\mathbf{S}}_r^{ - 1}{\mathbf{H}}} \right]^{ - 1}} \hfill \\
\end{split}
\end{equation}
where we denote ${{\mathbf{S}}_p}\left( {k|k - 1} \right)$ by ${{\mathbf{S}}_p}$, ${{\mathbf{S}}_r}\left( k \right)$ by ${{\mathbf{S}}_r}$, ${{\mathbf{C}}_x}\left( k \right)$ by ${{\mathbf{C}}_x}$ and ${{\mathbf{C}}_y}\left( k \right)$ by ${{\mathbf{C}}_y}$ for simplicity.
Using the matrix inversion lemma with the identification:
\begin{equation*}
\begin{gathered}
  {\left( {{\mathbf{S}}_p^{ - 1}} \right)^T}{{\mathbf{C}}_x}{\mathbf{S}}_p^{ - 1} \to {\mathbf{A}},{\text{ }}{{\mathbf{H}}^T} \to {\mathbf{B}}, \hfill \\
  {\mathbf{H}} \to {\mathbf{C}},{\text{ }}{\left( {{\mathbf{S}}_r^{ - 1}} \right)^T}{{\mathbf{C}}_y}{\mathbf{S}}_r^{ - 1} \to {\mathbf{D}}. \hfill \\
\end{gathered}
\end{equation*}
We arrive at
\begin{equation}\label{equ:a5}
\resizebox{1.0\hsize}{!}
{$\begin{split}
  & {\left( {{{\mathbf{W}}^T}\left( k \right){\mathbf{C}}\left( k \right){\mathbf{W}}\left( k \right)} \right)^{ - 1}} \hfill \\
   = & \left( {{{\mathbf{S}}_p}{\mathbf{C}}_x^{ - 1}{\mathbf{S}}_p^T - {{\mathbf{S}}_p}{\mathbf{C}}_x^{ - 1}{\mathbf{S}}_p^T{{\mathbf{H}}^T}{{({{\mathbf{S}}_r}{\mathbf{C}}_y^{ - 1}{\mathbf{S}}_r^T + {\mathbf{H}}{{\mathbf{S}}_p}{\mathbf{C}}_x^{ - 1}{\mathbf{S}}_p^T{{\mathbf{H}}^T})}^{ - 1}}{\mathbf{H}}{{\mathbf{S}}_p}{\mathbf{C}}_x^{ - 1}{\mathbf{S}}_p^T} \right) \hfill \\
\end{split}$}
\end{equation}
Furthermore, by (\ref{equ:a1}) $\sim$ (\ref{equ:a3}), we derive
\begin{equation}\label{equ:a6}
\resizebox{1.0\hsize}{!}
{$\begin{split}
&{{\mathbf{W}}^T}(k){\mathbf{C}}(k){\mathbf{D}}(k)\\
 = &{\left( {{\mathbf{S}}_p^{ - 1}} \right)^T}{{\mathbf{C}}_x}{\mathbf{S}}_p^{ - 1}\widehat {\mathbf{x}}(k|k - 1)\\
&+ {{\mathbf{H}}^T}{\left( {{\mathbf{S}}_r^{ - 1}} \right)^T}{{\mathbf{C}}_y}{\mathbf{S}}_r^{ - 1}\left( {{\mathbf{y}}(k) - {\mathop{\operatorname h}\nolimits} \left( {k,\widehat {\mathbf{x}}(k|k - 1)} \right) + {\mathbf{H}}(k)\widehat {\mathbf{x}}(k|k - 1)} \right)
\end{split}$}
\end{equation}
Combining (\ref{equ:32}), (\ref{equ:a5}) and (\ref{equ:a6}), we have (\ref{equ:33}).




\ifCLASSOPTIONcaptionsoff
  \newpage
\fi



\bibliographystyle{IEEEtran}
\bibliography{bare_jrnl}
\end{document}